\documentclass[conference]{IEEEtran}
\IEEEoverridecommandlockouts
\usepackage{cite}
\usepackage{amsmath,amssymb,amsfonts}
\usepackage{algorithmic}
\usepackage[ruled]{algorithm2e}
\usepackage{graphicx}
\usepackage{textcomp}
\usepackage{xcolor}
\usepackage{subfig}
\usepackage{booktabs}
\usepackage{multirow}
\usepackage{balance}

\newcommand{\ra}[1]{\renewcommand{\arraystretch}{#1}}
\def\BibTeX{{\rm B\kern-.05em{\sc i\kern-.025em b}\kern-.08em
    T\kern-.1667em\lower.7ex\hbox{E}\kern-.125emX}}
\begin{document}
\title{Temporal Convolutional Memory Networks for Remaining Useful Life Estimation of Industrial Machinery
}

\author{\IEEEauthorblockN{Lahiru Jayasinghe\IEEEauthorrefmark{1},
		Tharaka Samarasinghe\IEEEauthorrefmark{2},
		Chau Yuen\IEEEauthorrefmark{1},
		Jenny Chen Ni Low\IEEEauthorrefmark{4},
		Shuzhi Sam Ge\IEEEauthorrefmark{3}}
	\IEEEauthorblockA{SUTD-MIT International Design Centre, Singapore University of Technology and Design, Singapore.\IEEEauthorrefmark{1}\\
		Department of Electronic and Telecommunication Engineering, University of Moratuwa, Sri Lanka.\IEEEauthorrefmark{2}\\
		Department of Electrical and Electronic Engineering, University of Melbourne, Australia.\IEEEauthorrefmark{2}\\
		Keysight Technologies, Singapore.\IEEEauthorrefmark{4}\\
		Department of Electrical and Computer Engineering, National University of Singapore, Singapore.\IEEEauthorrefmark{3}\\
		Email: \{aruna$ \_ $jayasinghe,  yuenchau\}@sutd.edu.sg\IEEEauthorrefmark{1}, 
		tharakas@uom.lk\IEEEauthorrefmark{2},
		jenny-cn\_low@keysight.com\IEEEauthorrefmark{4},
		samge@nus.edu.sg\IEEEauthorrefmark{3}}}

\bibliographystyle{ieeetr}
\maketitle

\begin{abstract}
	Accurately estimating the remaining useful life (RUL) of industrial machinery is beneficial in many real-world applications. Estimation techniques have mainly utilized linear models or neural network based approaches with a focus on short term time dependencies. This paper, introduces a system model that incorporates temporal convolutions with both long term and  short term time dependencies. The proposed network  learns salient features and complex temporal variations in sensor values, and predicts the RUL. A data augmentation method is used for increased accuracy. The proposed method is compared with several state-of-the-art algorithms on publicly available datasets. It demonstrates promising results, with superior results for datasets obtained from complex environments.
\end{abstract}


\section{Introduction}
\let\thefootnote\relax\footnote{This work was supported in part by Keysight Technologies, International Design Center, and NSFC 61750110529.}
Accurate RUL estimation is crucial in prognostics and health management of industrial machinery such as aeroplanes, heavy vehicles and turbines. A system that predicts failures beforehand enables owners in making informed maintenance decisions, in advance, to prevent permanent damages. This leads to a significant reduction in operational and maintenance costs. Hence, RUL estimation is considered vital in industry operational research. Approaches for RUL estimation are mainly two-fold, and can be categorized as model-based or data-driven approaches \cite{results}. The Model-based approaches need a physical failure model to estimate the RUL, which in most practical scenarios, is difficult to produce. This limitation has  promoted data-driven approaches for RUL estimation. This paper, focuses on the implementation of a data-driven approach to estimate the RUL of industrial machinery.

The literature on RUL estimation includes sliding window approaches \cite{rulann}, hidden Markov model (HMM) based approaches \cite{rulhmm} and recurrent neural network (RNN) based approaches \cite{RUL}. The sliding window approach in \cite{rulann} only considers the relations within the sliding window, and hence, only short term time dependencies are captured. In HMM based approaches, where the hidden states only depend on the previous state,  modeling long time dependencies leads to high computational complexity and storage requirements. On the other hand, RNNs are capable of learning time dependencies more than HMMs, but face the vanishing gradient problem when used to capture long-term time dependencies \cite{vanishing}. 

Recently, convolutional neural networks (CNN) and long short-term memory (LSTM) networks  have emerged as efficient methods in many pattern recognition application domains such as computer vision \cite{vision,toilet}, surveillance \cite{drone,conf_icit}, and medicine \cite{survey_med}. Since the RUL estimation problem is closely related to pattern recognition, similar techniques can be applied to solve the RUL estimation problem as well \cite{dl_phm}. To this end, authors in \cite{cnn_rul} have proposed a 2D convolutional approach by using sliding windows for RUL estimation, and \cite{results} has proposed LSTM networks for RUL estimation. When comparing the two, even though LSTM networks are capable of building long term time dependencies, its feature extraction capabilities are marginally lower than CNN \cite{temporal_convolutions}. However, CNN and LSTM networks both possess unique abilities to learn features from data, and hence, we have utilized both techniques for RUL estimation in this paper.

Depending on the kernel size, 2D convolutions consider values from several sensors simultaneously when extracting features. This sometimes induces noise in the result. In contrast, 1D convolutions only occur in the temporal dimension of the given sensor, and they extract features without any interference from the other sensor values \cite{temporal_convolutions}. Therefore, we have used 1D temporal convolutions to learn features relevant to time dependencies of sensor values. The extracted features from the convolutions are then fed to a stacked LSTM network to learn the long short-term time dependencies. The paper also proposes an augmentation algorithm for the training stage to enhance the performance of the estimation. Paper validates and benchmarks the proposed system architecture with several state-of-the-art algorithms, using publicly available datasets.  The results are promising for all datasets. However, the specialty is that the architecture provides superior results for datasets obtained from complex environments, and this can be highlighted as the main contribution of the paper. 

The rest of the paper is organized as follows. Section \ref{problem_formulation} sets up the problem formulation and Section \ref{system_model} incrementally describes the whole system architecture. Section \ref{results} presents the results of the paper, and Section \ref{conclution} concludes the paper.

\section{Problem Formulation}\label{problem_formulation}

Consider a system with $N$ components ({\em e.g.} engines). $M$ sensors are installed on each component. The usual setting is to utilize vibration and temperature sensors to collect information about the machine's behaviour. The data from the $n$-th component throughout its useful lifetime produces a multivariate time series $X_n \in \mathbb{R}^{T_n \times M}$, where $T_n$ denotes the total number of time steps of component $n$ throughout its lifetime. $X_n$ is also known as the training trajectory of the $n$-th component.  $X_n^t \in \mathbb{R}^{M}$ denotes the $t$-th time step of $X_n$, where $t \in \{1, ..., T_n\}$, and it is a vector of $M$ sensor values. Hence, the training set is given by $\mathcal{X}=\{X_n^t | n = 1,\ldots, N; t = 1, \ldots, T_n\}$.

The RUL estimation is done based on test data. Test data is from a similar environment with $K$ components. $K$ may not be necessarily equal to $N$. The test data of the $k$-th component produces another multivariate time series $Z_k \in \mathbb{R}^{L_k \times M}$,  where $L_k$ denotes the total number of time steps related to component $k$ in the test data. $Z_k$ is also known as the test trajectory of the $k$-th component. The test set is given by $ \mathcal{Z}=\{Z_k^t | k = 1, \ldots, K; t = 1, \ldots,L_k \}$.  Obviously, the test set will not consist all the time steps up to the failure point, that is, $L_k$ will generally be smaller compared to the number of time steps taken for the failure of component $k$, which we denote by $\bar{L}_k$.  We focus on estimating the RUL of component $k$, which is given by $\bar{L}_k - L_k$, by utilizing time steps from 1 to $L_k$, that are included in the test set.

\subsection{Datasets}\label{subsection:dataset}

Publicly available NASA Commercial Modular Aero-Propulsion System Simulation dataset (C-MAPSS) \cite{CMAPSS}  is chosen for the benchmarking purposes, as it has been widely used in the literature. As given in Table \ref{tb:dataset}, C-MAPSS simulated dataset consists of 4 sub-datasets, with different operating and fault conditions, leading to complex relations with sensors. 

\begin{table}[]
	\caption{C-MAPSS Data Set \cite{CMAPSS}}
	\centering
	\label{tb:dataset}
	\resizebox{0.4\textwidth}{!}{
		\begin{tabular}{@{}lcccc@{}}
			\toprule
			Dataset              & FD001 & FD002 & FD003 & FD004 \\ \midrule
			Training trajectories   & 100   & 260   & 100   & 249   \\ \midrule
			Testing trajectories    & 100   & 259   & 100   & 248   \\ \midrule
			Operating conditions & 1     & 6     & 1     & 6     \\ \midrule
			Fault conditions     & 1     & 1     & 2     & 2     \\ \bottomrule
	\end{tabular}}
\end{table}

As shown in Table \ref{tb:dataset}, different sub-datasets (FD001, FD002, etc.) contain different number of training, and testing trajectories. The complexity of sub-datasets will increase with the number of operating conditions and fault conditions \cite{reveiw_CMAPSS}. Hence, FD002 and FD004 sub-datasets are considered to be the complex datasets. In every sub-dataset, training trajectories are concatenated along the temporal axis, and same applies for the testing trajectories as well. In general, these concatenated trajectories are included in a $l$-by-26 matrix, where $l$ denotes the total length after concatenation of the trajectories.  

In this $l$-by-26 matrix, the first column represents the engine ID, second column represents the operational cycle number, third to fifth columns represent the three operating settings that have a substantial effect on the engine performance \cite{cmapss_details}, and the last 21 columns represent the sensor values, {\em i.e.}, $M=21$. More information about the sensors are available in \cite{sensors}. The actual RUL values are provided to the dataset separately for verification purposes.

\subsection{Performance Evaluation} \label{evaluation}

In order to measure the performance of the RUL estimation, we use the \textit{Scoring Function} and the \textit{Root Mean Square Error} (RMSE). To this end, the error in estimating the RUL of the $n$-th component is given by
\begin{equation}\label{eq:error} E_n = RUL_{Estimated} - RUL_{True}.\end{equation}
It is not hard to see that $E_n$ can be both positive and negative. However, $E$ being positive will be more harmful, since the machine will fail before the estimated time. Therefore, a scoring function that penalizes positive $E_n$ value is used, and is given by
\begin{equation}\label{eq:score_function}S=\begin{cases} 
\sum_{i=1}^{N}(e^{-\frac{E_i}{13}}) & E_i< 0 \\
\sum_{i=1}^{N}(e^{-\frac{E_i}{10}}) & E_i\geq 0 
\end{cases}. \end{equation}  

\begin{figure}[!]
	\centering
	{\includegraphics[width=0.4\textwidth]{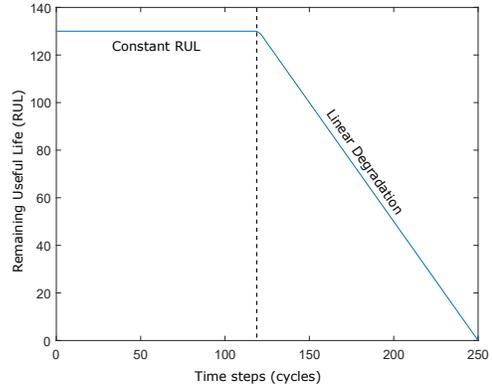}}
	\caption{Piece-wise linear RUL target function.}
	\label{fig:RUL}
\end{figure}

A main drawback of this scoring function is its sensitivity to outliers.  Since there is no error normalization and it follows an exponential curve, one single outlier can drastically change the score value. Therefore, we also use RMSE, which is given by   
\begin{equation}\label{eq:RMSE}
RMSE = \sqrt{\frac{1}{N}\sum_{i=1}^{N} E_i^2}.
\end{equation}

\subsection{RUL Target Function} \label{rul}

In this paper, we use the piecewise linear degradation model \cite{results} depicted in Fig. \ref{fig:RUL} as the target function in the estimation process. The degradation of the system typically starts after a certain degree of usage, and hence, we consider this model to be more suited compared to the linear degradation model \cite{RUL_linear}. The target function also has an upper bound for the maximum RUL, which avoids over estimations. Let $R_n \in \mathbb{R}^{T_n \times 1}$ represent the generated RUL values for component $n \in \{1, \ldots,N \}$ by utilizing the target function and the training trajectory $X_n$. Similar to the notations defined earlier, $R_n^t \in \mathbb{R}$ denotes the RUL value of component $n$ at the $t$-th time step. The RUL values for all $N$ training components (or trajectories) can be represented using the set  $\mathcal{R} = \{R_n^t|n=1,...,N;t=1,...,T_n\}$. These RUL values act as the labels for the supervise training in the proposed system architecture.

\section{System Architecture}\label{system_model}

\begin{figure}[t]
	\centering
	{\includegraphics[width=0.5\textwidth]{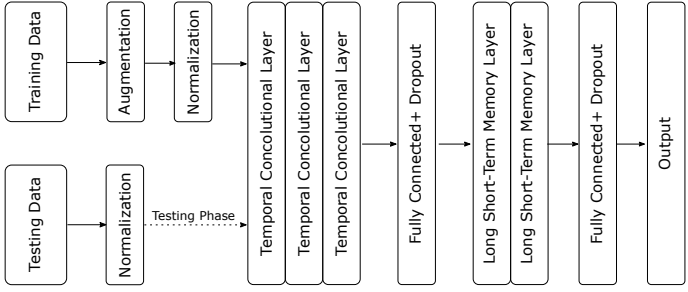}}
	\caption{The proposed system architecture for RUL estimation for C-MAPSS dataset.}
	\label{fig:system_model}
\end{figure}

The proposed system architecture consists of data preprocessing, data augmentation and a deep regression model for RUL estimation. As shown in Fig. \ref{fig:system_model}, both testing, and training data are normalized, but only the training data are augmented before feeding into the regression model. Stacked temporal convolution layers and LSTM layers, which are connected with each other using fully connected layers, have created the regression model for the proposed system architecture.

\begin{algorithm}[t]
	\KwIn{Training data set, $\mathcal{X} = \{X_n^t|n=1,...,N;t=1,...,T_n\}$ }
	\KwIn{Training RUL value set, $\mathcal{R} = \{R_n^t|n=1,...,N;t=1,...,T_n\}$ }
	\KwIn{Augmentation size, $\lambda \in \mathbb{Z}^+$} 
	
	\KwOut{Augmented, training dataset $\mathcal{X'}$ and its RUL value set $\mathcal{R'}$.}
	
	$\mathcal{X'} = \mathcal{X}$
	, $\mathcal{R'} = \mathcal{R}$\;
	\For{each component $n=1,...,N$}
	{
		\For {each time step $t=1 \ to \ T_n$}
		{
			\If{$R_n^t < R_n^{t-1}$}
			{
				\For{$i=1 \ to \ \lambda$}
				{
					$t_i = DiscreteUniformRND(t,T_n)$
					$\mathcal{X}_{n,i}=\{X_n^1, ..., X_n^{t_i}\}$ $\mathcal{R}_{n,i}=\{R_n^1, ..., R_n^{t_i}\}$
					
				}
				Break;
			}
			
		}
		$\mathcal{X'} =\mathcal{X'} \bigcup_{i=1}^{\lambda}\mathcal{\mathcal{X}}_{n,i}$ 
		, $\mathcal{R'} =\mathcal{R'} \bigcup_{i=1}^{\lambda}\mathcal{\mathcal{R}}_{n,i}$ \label{alg:sum}
	}
	\KwResult{$\mathcal{X'}$, $\mathcal{R'}$}
	\caption{Data Augmentation}
	\label{alg:1}
\end{algorithm}

\subsection{Data Normalization}\label{preprocessing}

According to the literature \cite{results}, data points can be clustered based on their respective operating conditions and normalization can be done based on those clusters. However, only FD002 and FD004 have six operating conditions, whereas FD001 and FD003 have only one operating condition, thus we have omitted the clustering. Alternatively, individual sensor values are normalized, and the value of the $i$-th sensor is normalized as
\begin{equation}\label{eg:normal}
\tilde{x_i} = \frac{x_i - \mu_{i}}{\sigma_i},
\end{equation} where $\mu_{i}$ and $\sigma_i$ denote the mean value and the standard deviation of the $i$-th sensor value, respectively.

\subsection{Data Augmentation}\label{aug}

\begin{figure}[t]
	\centering
	\subfloat[][Before augmentation]{
		\includegraphics[width=0.4\textwidth]{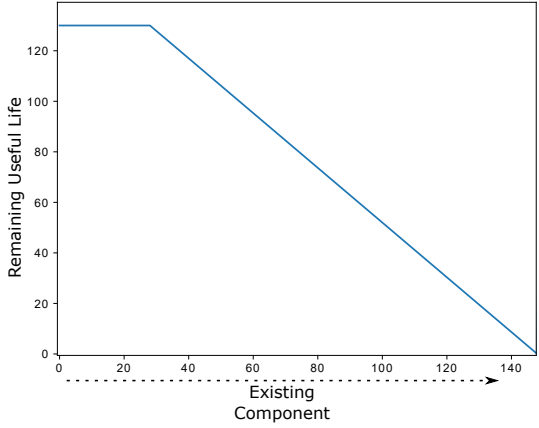}
		\label{fig:before_augmentation_fig}
	}
	
	\subfloat[][After augmentation]{
		\includegraphics[width=0.4\textwidth]{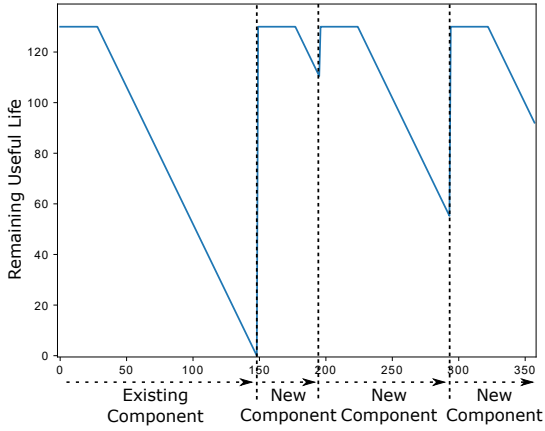}
		\label{fig:after_augmentation_fig}
	}
	\caption{
		Demonstrating the behaviour of the augmentation algorithm by using target functions from FD002 dataset.}
	\label{fig:aug}%
\end{figure}

\begin{figure}[t]
	\centering
	{\includegraphics[width=0.4\textwidth]{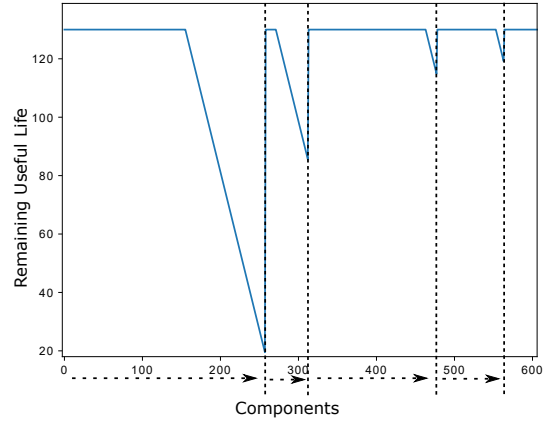}}
	\caption{Part of augmented training dataset.}
	\label{fig:testing_concat_rul}
\end{figure}

In this paper, a data augmentation algorithm is proposed to enhance the RUL estimation performance. Fig. \ref{fig:before_augmentation_fig} represents the target function of a complete training trajectory (before augmentation). In the augmentation algorithm, we have utilized the complete training trajectory to generate partial training trajectories, as illustrated in Fig. \ref{fig:after_augmentation_fig}. According to the example in Fig. \ref{fig:after_augmentation_fig}, we have generated three partial training trajectories using the complete training trajectory in  Fig. \ref{fig:before_augmentation_fig}. Each partial training trajectory is obtained by truncating the complete training trajectory at a random point along the linear degradation. Note that partial trajectories have a closer resemblance to test data, as they are truncated earlier to the failure point. It is well known that supervise algorithms perform well for patterns they have encountered previously in the training phase, and hence, this augmentation leads to better learning, and increases the accuracy of the estimation.  Fig. \ref{fig:testing_concat_rul} represents target functions of a part of the training dataset after data augmentation, and this training dataset is used for learning.

These ideas are formally presented through Algorithm \ref{alg:1}.  We have the training set $\mathcal{X}$, training RUL value set obtained from the target function $\mathcal{R}$, and an integer $\lambda$ as inputs, where $\lambda$ denotes the number of partial trajectories to be generated through data augmentation. For each training trajectory, the algorithm searches for the time step $t$, where the RUL value starts to decrease. This is done to capture the starting time step of the linear degradation. If such a time step exists for component $n \in \{1,\ldots,N \}$, then a random integer is drawn between $t$ and $T_n$ from a discrete uniform distribution, which is represented as $DiscreteUniformRND(t,T_n)$ in the algorithm. $\lambda$ random integers are generated for each training trajectory. If the $i$-th random integer for the $n$-th component is $t_i$, then sequences related to time steps $1$ to $t_i$ from $X_n$ and $R_n$  are selected as the partial training trajectories, {\em i.e.}, $\mathcal{X}_{n,i}$ and $\mathcal{R}_{n,i}$, respectively. Then, these sets are added to the original training dataset $\mathcal{X}$ and the RUL value set $\mathcal{R}$. Since this process is carried out $\lambda$ times for each component $n$, the cardinality of  the training set and the RUL value set will increase by $\lambda+1$ times, compared to its original size.

\subsection{Temporal Convolutional Layer}\label{cnn}

The temporal convolutional layer consists of 1D-convolution and 1D-max-pooling. With regards to temporal convolution, let $d^{(l-1)}$ and $d^{(l)}$ be the input and the output of the $l$-th layer, respectively. Input to the $l$-th layer is the output of the $(l-1)$-th layer. Since there are several feature maps for a layer, we denote the $j$-th feature map of layer $l$ as $d_j^{(l)}$, and this can be computed by \begin{equation}\label{eq:conv}
d_j^{(l)}=f\left( \sum_{i}d_i^{(l-1)}*\vec{w}_{i,j}^{(l)} + b_j^{(l)}\right), 
\end{equation} where $*$ denotes the convolution operator, $\vec{w}_{i,j}^{(l)}$, and $ b_j^{(l)}$ represent the 1-D weight kernel and the bias of the $j$-th feature map of the $l$-th layer, respectively, and $f$ is a non-linear activation function. Often, it would be a rectified linear unit activation (ReLu).  

The sub-sampling or pooling works as a progressive mechanism to reduce the spatial size of the feature representations. This increases computational efficiency and reduces parameters that control over-fitting of neurones. Max-pooling is the most accepted operation of sub-sampling in CNN. This can operate independently from the convolutional operation. The 1D-max-pooling is given by \begin{equation}\label{eq:max}
d_{j_i}^{(l)}= \max \left(d_{j_{i_{nbh}}}^{(l)}\right),
\end{equation} where $d_{j_i}^{(l)}$ denotes the $i$-th element of feature map $d_j^{(l)}$, $d_{j_{i_{nbh}}}^{(l)}$ denotes the set of values in the 1D-neighbourhood of $d_{j_i}^{(l)}$. The neighbourhood size is defined by the 1D-pooling size. 

\subsection{Long Short-Term Memory Layer}\label{lstm}    

For a given input sequence $X_n=\left( X_n^1,...,X_n^T \right) $, a recurrent neural network (RNN) generates an output of $Y_n=\left( Y_n^1,...,Y_n^T \right)$ using the hidden vector sequence of $h^t=\left( h^1,...,h^T \right)$ . This is done by iterating the following equation from $t=1$ to $t=T$: \begin{equation}
h^t=f \left(W_{xh}X_n^t+W_{hh}h^{t-1}+b_h  \right),
\end{equation}\begin{equation}
Y_n^t=W_{hy}h^t+b_y,
\end{equation} where $W_{xh}$, $W_{hh}$ and $W_{hy}$ denote transformation matrices of the input and hidden vector, and $b_h$, $b_y$ are the bias vectors. Although this RNN combine the temporal variations in the output, it lacks memory connectivities. Therefore, memory gates are introduced into the RNN cells, and they are known as  long short-term memory networks (LSTM)\cite{lstm_new}. The calculation of $h^t$ for the LSTM networks follows from \begin{equation}\label{eq:lstm}
i^t=\sigma\left(W_{ix}X_n^t+ W_{ih}h^{t-1}+W_{ic}c^{t-1}+b_i  \right),
\end{equation} \begin{equation}
f^t=\sigma\left(W_{fx}X_n^t+ W_{fh}h^{t-1}+W_{fc}c^{t-1}+b_f \right),
\end{equation} \begin{equation}
c^t=f^tc^{t-1}+i^t\tanh\left(W_{cx}X_n^t+ W_{ch}h^{t-1}+b_c \right),
\end{equation} \begin{equation}
o^t=\sigma\left(W_{ox}X_n^t+ W_{oh}h^{t-1}+W_{oc}c^{t}+b_o  \right),
\end{equation} \begin{equation}
h^t=o^t\tanh\left(c^t\right),
\end{equation} where $\sigma$ is the logistic sigmoid function, and $i, f, o $ and $c$ denote the input gate, forget gate, output gate and cell activation vectors, respectively. The transformation weights $W_{ix}$, $W_{ih}$, $W_{ic}$, $W_{fx}$, $W_{fh}$, $W_{fc}$, $W_{cx}$, $W_{ch}$, $W_{ox}$, $W_{oh}$, $W_{oc}$, and bias values $b_i$, $b_f$, $b_c$, $b_o$ are computed during the training process. The input gate $i^t$, output gate $o^t$, and forget gate $f^t$ control the information flow within the LSTM network. Since there are several gates, this network is capable of keeping selective memory compared to an RNN. An array of LSTM is known as an LSTM layer.

 \begin{table}[t]
	\caption{Layer Details of the System Architecture \\ \footnotesize($channels=24, sequ ence\_length=100$)}
	\label{tb:system_model}
	\ra{1.3}
	\resizebox{0.45\textwidth}{!}{
		\begin{tabular}{@{}cll@{}}
			\toprule
			\multirow{2}{*}{Layer 1} & 1D-convolution  & filters=18, kernel\_size=2, strides=1,padding=same, activation=ReLu \\ \cmidrule(l){2-3} 
			& 1D-max-pooling  & pool\_size=2, strides=2, padding=same                               \\ \midrule
			\multirow{2}{*}{Layer2}  & 1D-convolution  & filters=36, kernel\_size=2, strides=1,padding=same, activation=ReLu \\ \cmidrule(l){2-3} 
			& 1D-max-pooling  & pool\_size=2, strides=2, padding=same                               \\ \midrule
			\multirow{2}{*}{Layer3}  & 1D-convolution  & filters=72, kernel\_size=2, strides=1,padding=same, activation=ReLu \\ \cmidrule(l){2-3} 
			& 1D-max-pooling  & pool\_size=2, strides=2, padding=same                               \\ \midrule
			\multirow{2}{*}{Layer4}  & fully-connected & layer\_size=sequence\_length*channels, activation=ReLu           \\ \cmidrule(l){2-3} 
			& dropout         & dropout\_probability = 0.2                                           \\ \midrule
			\multirow{2}{*}{Layer5}  & LSTM            & units = channels*3                                                          \\ \cmidrule(l){2-3} 
			& dropout-wrapper & dropout probability = 0.2                                           \\ \midrule
			\multirow{2}{*}{Layer6}  & LSTM            & units = channels*3                                                    \\ \cmidrule(l){2-3} 
			& dropout-wrapper & dropout probability = 0.2                                           \\ \midrule
			\multirow{2}{*}{Layer7}  & fully-connected & layer\_size=50, activation=ReLu                                     \\ \cmidrule(l){2-3} 
			& dropout         & dropout probability = 0.2                                           \\ \midrule
			Layer8                   & fully-connected & layer\_size=1 (output layer)                                        \\ \bottomrule
	\end{tabular}}
\end{table}

\subsection{Temporal Convolutional Memory Networks}\label{cnn_lstm} 

The proposed architecture is implemented by combining temporal convolutional layers with LSTM layers through a fully-connected layer. Let $d_{j}^{flat}$, where $j \in \{1,...,Q\}$, be the $j$-th flattened feature map of the last temporal convolution layer. The set of flattened feature maps in the last layer can be represented as $D^{flat}$ = $\{d_j^{flat}|j=1,...,Q\} $. This flattened layer passes through a fully-connected neural network as follows: \begin{equation}
h_{fc} = f\left( W_{fc}D^{flat}+ b_{fc}\right),
\end{equation} where $h_{fc}$ denotes the output of the fully-connected layer, and $W_{fc}$ and $b_{fc}$ denote the transformation weights and bias of the fully-connected layer.

Proposed system architecture consists of three temporal convolutional layers. Out of them, the first layer consist of 18 filters, the second layer consists of 36 filters, and the final convolution layer consist of 72 filters. As shown in the Table. \ref{tb:system_model}, every convolution layer is followed by a 1D-max-pooling layer, and size two kernels are used in every convolution and pooling operation. A fully-connected layer with drop-out regularization has been introduced to connect the temporal convolutional layer to the LSTM layers. As described in Subsection \ref{subsection:dataset}, data contains 3 operating settings and 21 sensor readings. Hence, the number of channels in the input equals to 24, and the sequence length equals to 100. Therefore, the input to the first temporal convolutional layer can be represented as $\{X_n^t|t=t_i,...,t_i+100\}$, where $t_i$ is the end time step of the previous input of the first temporal convolutional layer. The LSTM layer size has been decided empirically. The final fully-connected layers work as regression layers to estimate the RUL values.

\section{Results and Discussion}\label{results}
We have performed extensive experiments to evaluate the proposed system architecture. This section discusses the performance of the augmentation algorithm, the impact of the CNN and LSTM on RUL estimation, and the RUL evaluation and benchmarking results.

\begin{table}[]
	\caption{Performances Analysis of the Augmentation Algorithm for the System Architecture. \\ {\footnotesize$Performance \ Gain =\left( 1-\frac{Including Augmentation}{Excluding Augmentation} \right)  \times 100\%  $} }
	\label{tb:aug}
	\resizebox{0.5\textwidth}{!}{
		\begin{tabular}{@{}ccc|cc|cc|cc@{}}
			\toprule
			Dataset                                                                       & \multicolumn{2}{c}{FD001}                                 & \multicolumn{2}{c}{FD002}                                 & \multicolumn{2}{c}{FD003}                                 & \multicolumn{2}{c}{FD004}                                 \\ \midrule
			Evaluation                                                                    & Score                                             & RMSE  & Score                                             & RMSE  & Score                                             & RMSE  & Score                                             & RMSE  \\ \midrule
			\begin{tabular}[c]{@{}c@{}}Excluding\\ Augmentation\end{tabular} & \begin{tabular}[c]{@{}c@{}}1.67\\ $\times10^4$\end{tabular} & 31.60 & \begin{tabular}[c]{@{}c@{}}3.55\\ $\times10^4$\end{tabular} & 31.06 & \begin{tabular}[c]{@{}c@{}}1.18\\ $\times10^4$\end{tabular} & 38.25 & \begin{tabular}[c]{@{}c@{}}5.91\\ $\times10^4$\end{tabular} & 36.85 \\ \midrule
			\begin{tabular}[c]{@{}c@{}}Including\\ Augmentation\end{tabular}    & \begin{tabular}[c]{@{}c@{}}2.41\\ $\times10^3$\end{tabular} & 29.55 & \begin{tabular}[c]{@{}c@{}}4.19\\ $\times10^3$\end{tabular} & 21.03 & \begin{tabular}[c]{@{}c@{}}3.44\\ $\times10^3$\end{tabular} & 27.11 & \begin{tabular}[c]{@{}c@{}}7.96\\ $\times10^3$\end{tabular} & 23.57 \\ \midrule
			\begin{tabular}[c]{@{}c@{}}Performance \\ Gain \end{tabular}    & \begin{tabular}[c]{@{}c@{}}\textbf{85.56}\end{tabular} & \textbf{6.48} & \begin{tabular}[c]{@{}c@{}}\textbf{88.19}\end{tabular} & \textbf{40.76} & \begin{tabular}[c]{@{}c@{}}\textbf{70.84}\end{tabular} & \textbf{29.12} & \begin{tabular}[c]{@{}c@{}}\textbf{86.53}\end{tabular} & \textbf{36.03} \\ \bottomrule
	\end{tabular}}
\end{table}

\paragraph{Augmentation} We trained our model, including and excluding the augmentation step for the same number of training iterations, with the same hyper-parameters. As shown in Table \ref{tb:aug}, the augmentation improved the performance drastically, specially in FD002 and FD004.

\begin{table}[]
	\caption{Evaluation and Benchmarking Results}
	\label{tb:results}
	\resizebox{0.5\textwidth}{!}{
		\begin{tabular}{@{}ccc|cc|cc|cc@{}}
			\toprule
			Dataset                                                                       & \multicolumn{2}{c}{FD001}                                 & \multicolumn{2}{c}{FD002}                                 & \multicolumn{2}{c}{FD003}                                 & \multicolumn{2}{c}{FD004}                                 \\ \midrule
			Evaluation                                                                    & Score                                             & RMSE  & Score                                             & RMSE  & Score                                             & RMSE  & Score                                             & RMSE  \\ \midrule
			\begin{tabular}[c]{@{}c@{}}MLP \cite{results}\end{tabular} & \begin{tabular}[c]{@{}c@{}}1.80 $\times10^4$\end{tabular} & 37.56 & \begin{tabular}[c]{@{}c@{}}7.80 $\times10^6$\end{tabular} & 80.03 & \begin{tabular}[c]{@{}c@{}}1.74 $\times10^4$\end{tabular} & 37.39 & \begin{tabular}[c]{@{}c@{}}5.62 $\times10^6$\end{tabular} & 77.37 \\ \midrule
			\begin{tabular}[c]{@{}c@{}}SVR \cite{results}\end{tabular} & \begin{tabular}[c]{@{}c@{}}1.38 $\times10^3$\end{tabular} & 20.96 & \begin{tabular}[c]{@{}c@{}}5.90 $\times10^5$\end{tabular} & 42.00 & \begin{tabular}[c]{@{}c@{}}1.60 $\times10^3$\end{tabular} & 21.05 & \begin{tabular}[c]{@{}c@{}}3.71 $\times10^5$\end{tabular} & 45.35 \\ \midrule
			\begin{tabular}[c]{@{}c@{}}RVR \cite{results}\end{tabular} & \begin{tabular}[c]{@{}c@{}}1.50 $\times10^3$\end{tabular} & 23.80 & \begin{tabular}[c]{@{}c@{}}1.74 $\times10^4$\end{tabular} & 31.30 & \begin{tabular}[c]{@{}c@{}}1.43 $\times10^3$\end{tabular} & 22.37 & \begin{tabular}[c]{@{}c@{}}2.65 $\times10^4$\end{tabular} & 34.34 \\ \midrule
			\begin{tabular}[c]{@{}c@{}}CNN \cite{results}\end{tabular} & \begin{tabular}[c]{@{}c@{}}1.29 $\times10^3$\end{tabular} & 18.45 & \begin{tabular}[c]{@{}c@{}}1.36 $\times10^4$\end{tabular} & 30.29 & \begin{tabular}[c]{@{}c@{}}1.60 $\times10^3$\end{tabular} & 19.82 & \begin{tabular}[c]{@{}c@{}}5.55 $\times10^3$\end{tabular} & 29.16 \\ \midrule
			\begin{tabular}[c]{@{}c@{}}LSTM \cite{results}\end{tabular} & \begin{tabular}[c]{@{}c@{}}\textbf{3.38} $\mathbf{\times10^2}$\end{tabular} & \textbf{16.14} & \begin{tabular}[c]{@{}c@{}}4.45 $\times10^3$\end{tabular} & 24.49 & \begin{tabular}[c]{@{}c@{}}\textbf{8.52} $\mathbf{\times10^2}$\end{tabular} & \textbf{16.18} & \begin{tabular}[c]{@{}c@{}}5.55 $\times10^3$\end{tabular} & 28.17 \\ \midrule
			\begin{tabular}[c]{@{}c@{}}Proposed\\ Architecture\end{tabular}    & \begin{tabular}[c]{@{}c@{}}1.22 $\times10^3$\end{tabular} & 23.57 & \begin{tabular}[c]{@{}c@{}}\textbf{3.10} $\mathbf{\times10^3}$\end{tabular} & \textbf{20.45} & \begin{tabular}[c]{@{}c@{}}1.30 $\times10^3$\end{tabular} & 21.17 & \begin{tabular}[c]{@{}c@{}}\textbf{4.00} $\mathbf{\times10^3}$\end{tabular} & \textbf{21.03} \\ \bottomrule
	\end{tabular}}
\end{table}

\begin{figure*}[t]
	\centering%
	\subfloat[][System architecture without LSTM layers]{%
		\includegraphics[width=0.32\textwidth]{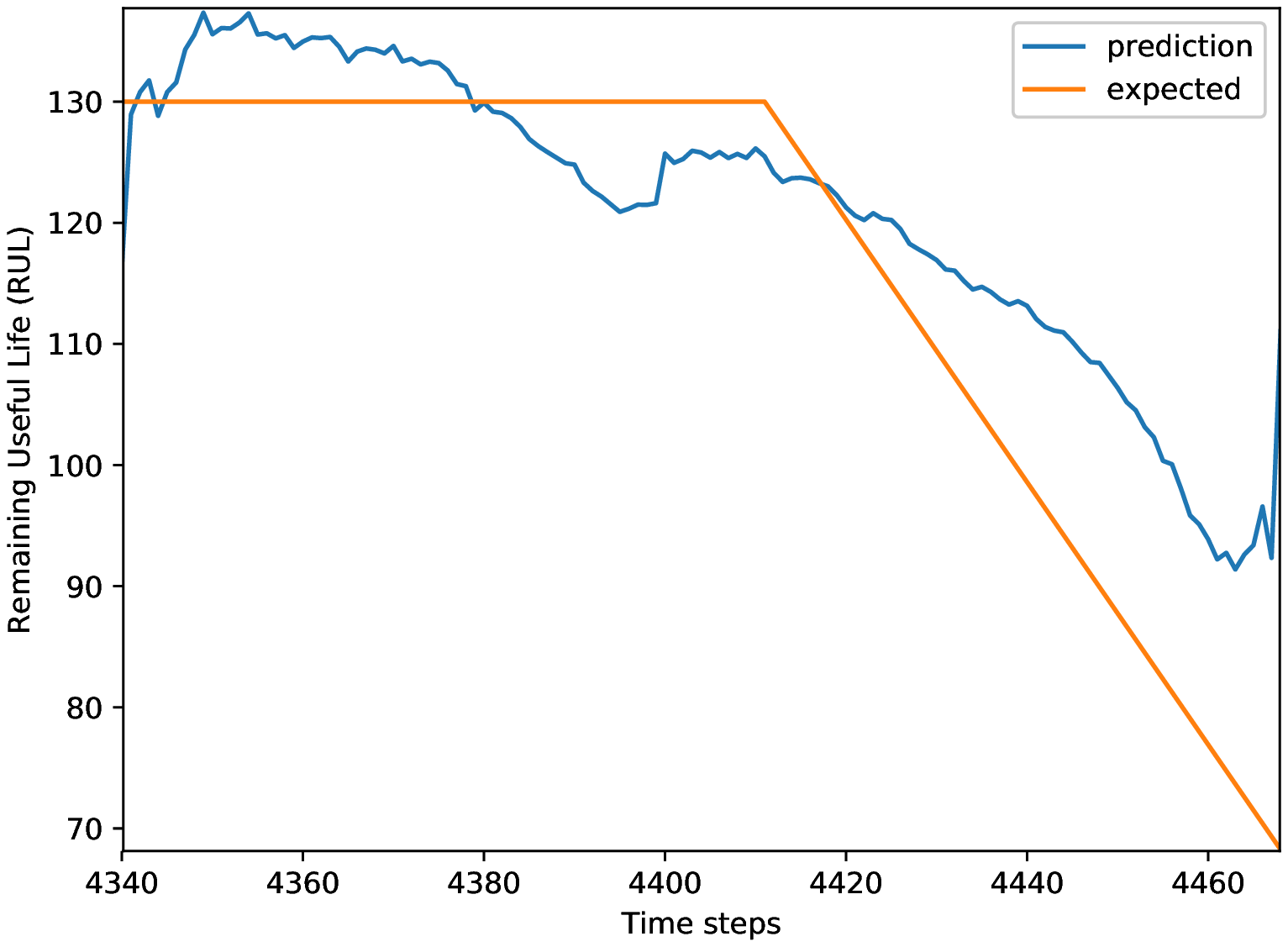}
		\label{fig:cnn}%
	}
	\subfloat[][System architecture without temporal convolutions]{%
		\includegraphics[width=0.32\textwidth]{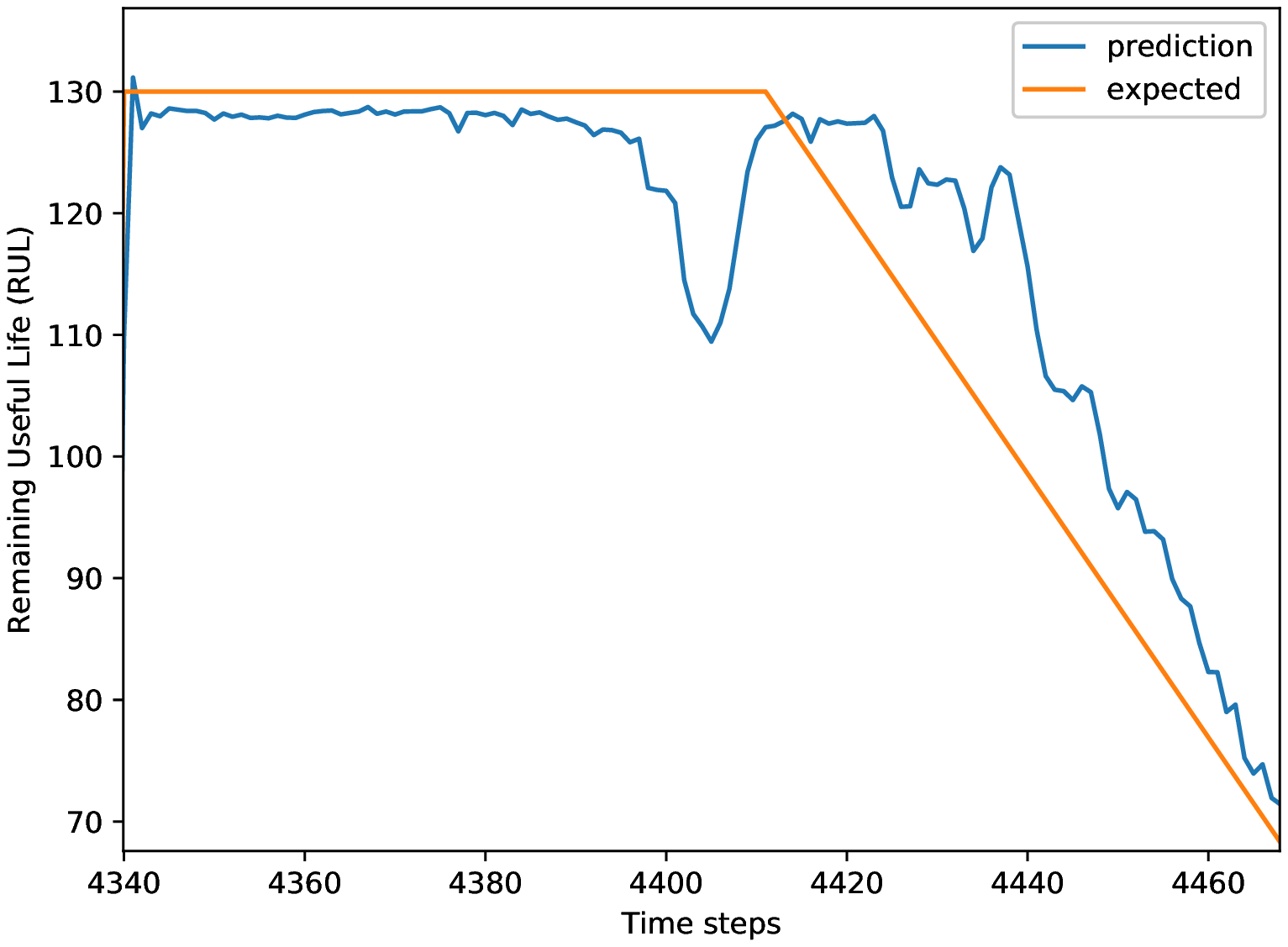}
		\label{fig:rnn}%
	}%
	\subfloat[][Proposed system architecture]{%
		\includegraphics[width=0.32\textwidth]{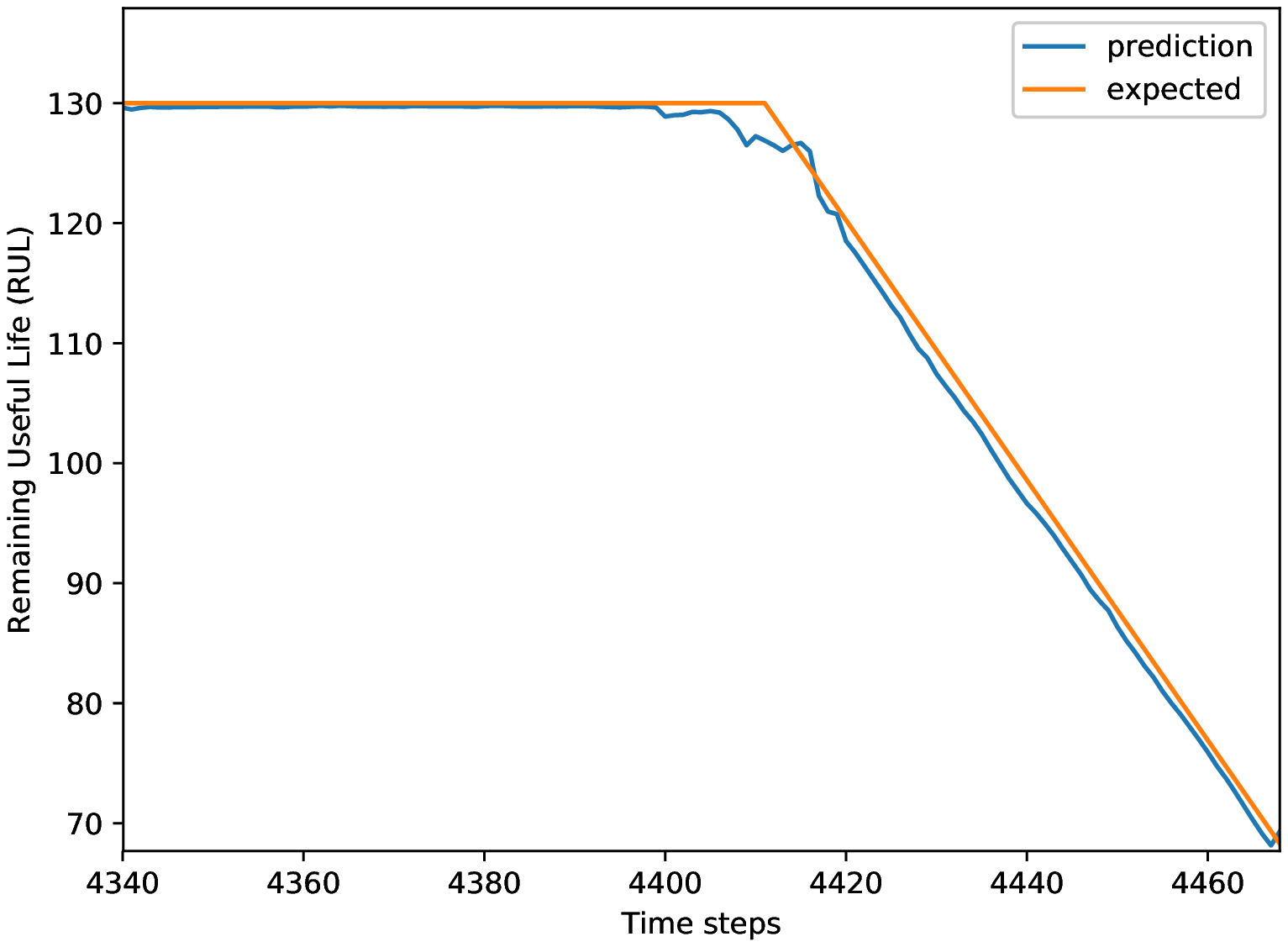}
		\label{fig:rulnet}%
	}%
	\caption{
		Results illustrating the effect of LSTM layers and temporal convolution layers in the proposed system architecture}%
	\label{fig:model_comparison}%
\end{figure*}

\paragraph{Impact of CNN and LSTM} Then we compared the proposed model with a scenario where the LSTM layers were omitted, and then with a scenario where the temporal convolution layers were omitted, while keeping the hyper-parameters the same. As shown in Fig. \ref {fig:model_comparison}, the system architecture with both temporal convolutions and LSTM follows the target function much accurately.

\paragraph{Evaluation Results} Past studies have shown that the estimated accuracies of FD002 and FD004 sub-datasets are low compared to FD001 and FD003, see Table. \ref{tb:results}. Even though the FD002 and FD004 datasets have six operating conditions and the complex relations among sensors, the proposed system architecture achieved the lowest RMSE values and score values by surpassing these issues. Since we treated all sub-datasets equally without clustering as explained in Subsection \ref{preprocessing}, RMSE values for all datasets are nearly equal. The reason behind score values being different is that they follow an exponential curve as given in (\ref{eq:score_function}). This observation implies that the proposed system architecture is capable of achieving better results, and is more robust to dataset complexities.  

\section{Conclusions and Future Work}\label{conclution}

This paper has presented a novel system architecture to estimate the RUL of an industrial machine. The proposed method outperforms previous studies specially in cases where the datasets are obtained from complex environments. The performance of the proposed system architecture mainly depends on the combination of temporal convolution layers and LSTM layers with data augmentation. Open areas to examine include on-line learning of data, and the performance of newer deep architectures on RUL estimation. Potential future work also include improving the performances on FD001 and FD003 datasets, and evaluating the performance on other publicly available datasets, for further improvements.

\balance
\footnotesize {\bibliography{bibfile}}

\end{document}